\documentclass[10pt,twocolumn,letterpaper]{article}

\usepackage{cvpr}
\usepackage{times}
\usepackage{epsfig}
\usepackage{graphicx}
\usepackage{amsmath}
\usepackage{amssymb}
\usepackage{cite}
\usepackage{subfigure}
\usepackage{dsfont}
\usepackage{multirow}

\usepackage[breaklinks=true,bookmarks=false]{hyperref}

\cvprfinalcopy 

\def\cvprPaperID{554} 

\begin{document}

\title{Social Behavior Prediction from First Person Videos}

\author{Shan Su\\
UPenn\\
{\tt\small sushan@seas.upenn.edu}
\and
Jung Pyo Hong\\
KAIST\\
{\tt\small jphong@ai.kaist.ac.kr}
\and
Jianbo Shi\\
UPenn\\
{\tt\small jshi@seas.upenn.edu}
\and
Hyun Soo Park\\
UMN\\
{\tt\small hspark@umn.edu}
}

\maketitle

\begin{abstract}
This paper presents a method to predict the future movements (location and gaze direction) of basketball players as a whole from their first person videos. The predicted behaviors reflect an individual physical space that affords to take the next actions while conforming to social behaviors by engaging to joint attention. Our key innovation is to use the 3D reconstruction of multiple first person cameras to automatically annotate each other's the visual semantics of social configurations.

We leverage two learning signals uniquely embedded in first person videos. Individually, a first person video records the visual semantics of a spatial and social layout around a person that allows associating with past similar situations. Collectively, first person videos follow joint attention that can link the individuals to a group. 
We learn the egocentric visual semantics of group movements using a Siamese neural network to retrieve future trajectories. We consolidate the retrieved trajectories from all players by maximizing a measure of social compatibility---the gaze alignment towards joint attention predicted by their social formation, where the dynamics of joint attention is learned by a long-term recurrent convolutional network. This allows us to characterize which social configuration is more plausible and predict future group trajectories.
\end{abstract}

\section{Introduction}
We {\em physically} interact with people around us while {\em mentally} engaging with them via joint attention. For example, you as an audience in a concert are locally affected by the people around you and are globally connected to the people on the other side of the stage by sharing joint attention. While the physical connection delineates the proximal space around us, the mental connection encodes the group's intent in a way that facilitates communications, role playing, and group task accomplishment. These connections provide social cues to further reason about the spatial and temporal extent of the social behaviors, which is a key design factor for an artificial intelligence of social robots.  

\begin{figure}[t]
  \centering  
  \label{Fig:teaser}\includegraphics[trim={0 0 0 60mm},clip,width=0.5\textwidth]{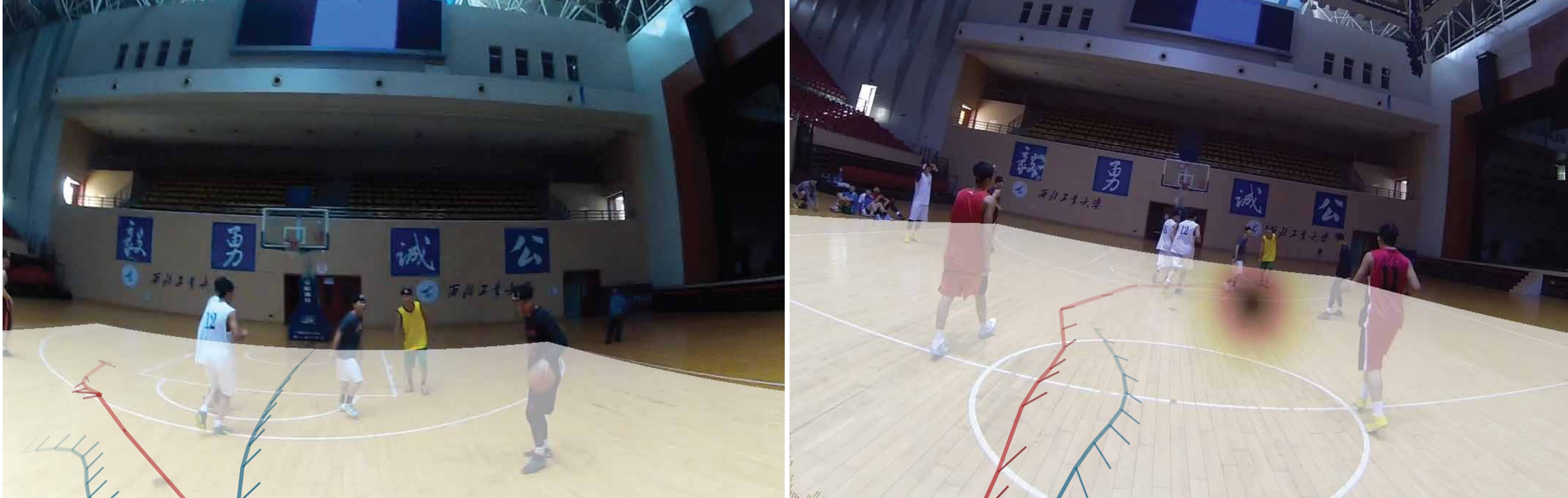}
  \caption{We predict a group trajectory of basketball players from first person videos. The red is the ground truth and blue is the predicted trajectories with gaze direction.} 
\end{figure}

However, such social cues are rather ambiguous, subtle, and situation dependent, which is challenging to be computationally learned by third person computer vision systems~\cite{kim:2010, kitani:2012,alahi:2016,ma:2016} due to their limited expressibility: it is necessary to tap into what we actually see. In this paper, we propose to use first person cameras collectively to decode the social cues and to further predict their future social behaviors.

What visual information makes us to stay connected to people, physically and mentally? We conjecture that two unique signals recorded in first person videos can describe the connections. 
(1) Individually, a first person video encodes the egocentric visual semantics that provides a social and spatial context to take the next action. (2) Collectively, first person videos follow joint attention spatially arranged by social formation~\cite{kendon:1990,park:2015}, e.g., audiences dynamically change their social formation to secure visibility, which links the individuals to a group. As a proof-of-concept, we integrate these two learning signals to predict the movement (location and gaze directions) of basketball players, one of most complex forms of social interactions, from their first person videos (Figure~\ref{Fig:teaser}).  

Our method takes an input, the first person videos of basketball players and outputs a set of plausible future trajectories. We learn an egocentric visual representation to recognize similar social and spatial configurations, e.g., which makes us to move, using a Siamese neural network. This representation is used to retrieve a set of future trajectories per player. We find a plausible group trajectory set from the retrieved trajectories of all players by maximizing a measure of social compatibility---the gaze alignment towards joint attention predicted by their social formation---via a generalized Dijkstra algorithm. The dynamics of joint attention is learned by a long-term recurrent convolutional network (LRCN) based on social formation features that encode locations and velocities of the players. Note that we predict not only the future locations but also their gaze directions and joint attention. 


Our key innovation is leveraging 3D reconstruction of multiple first person videos to automatically annotate each other's visual semantics of social configurations. This labels the location, orientation, and velocity of other players in pixels, precisely (reprojection error is often less than 0.5 pixel). This makes learning visual social signals on a large scale possible, which provides a richer context of the interactions comparing to third person social activity predictions~\cite{kim:2010, kitani:2012,alahi:2016,ma:2016}.



A challenge of using first person cameras is that they often produce highly jittery, blurry, and narrow view, unlike third person videos captured from mostly static and omniscient views. We virtually stabilize first person images by applying cylindrical projection, and directly learn visual semantics of social configurations from the images via a convolutional neural network. To resolve a limited visibility issue, we consolidate first person images of all players, which substantially extend visible space via 3D registration.

The first person videos have been increasingly adapted to record professional sports such as basketball, soccer, handball, ice hockey, and American football~\cite{firstvision}. Our work provides a computational tool to measure team performance and train players based on how they interact with others based on what they see. Beyond sports, decoding such social sensorimotor behaviors can be used to further explain how social cues are encoded in the human mirror neural system~\cite{rizzolatti:2004}. Also this social intelligence system can apply to content generation for social virtual/augmented reality~\cite{social_vr}, human-robot interactions, and collaborative education.





\noindent\textbf{Contribution} To our best knowledge, this is the first paper that predicts long-term activities from a collection of first person videos. The core technical contributions include (a) learning egocentric visual semantics to recognize social and spatial configurations, (b) using a measure of social compatibility to identify plausibility of social behaviors, (c) formulating the trajectory selection process using a dynamic programming, and (d) learning the dynamics of joint attention via LSTM. We demonstrate the predictive validity of our algorithm in real world basketball datasets by comparing with third person prediction systems. 

\section{Related Work}
\begin{figure*}[th]
  \centering  
      \subfigure[Geometry]{\label{Fig:concept}\includegraphics[height=0.137\textheight]{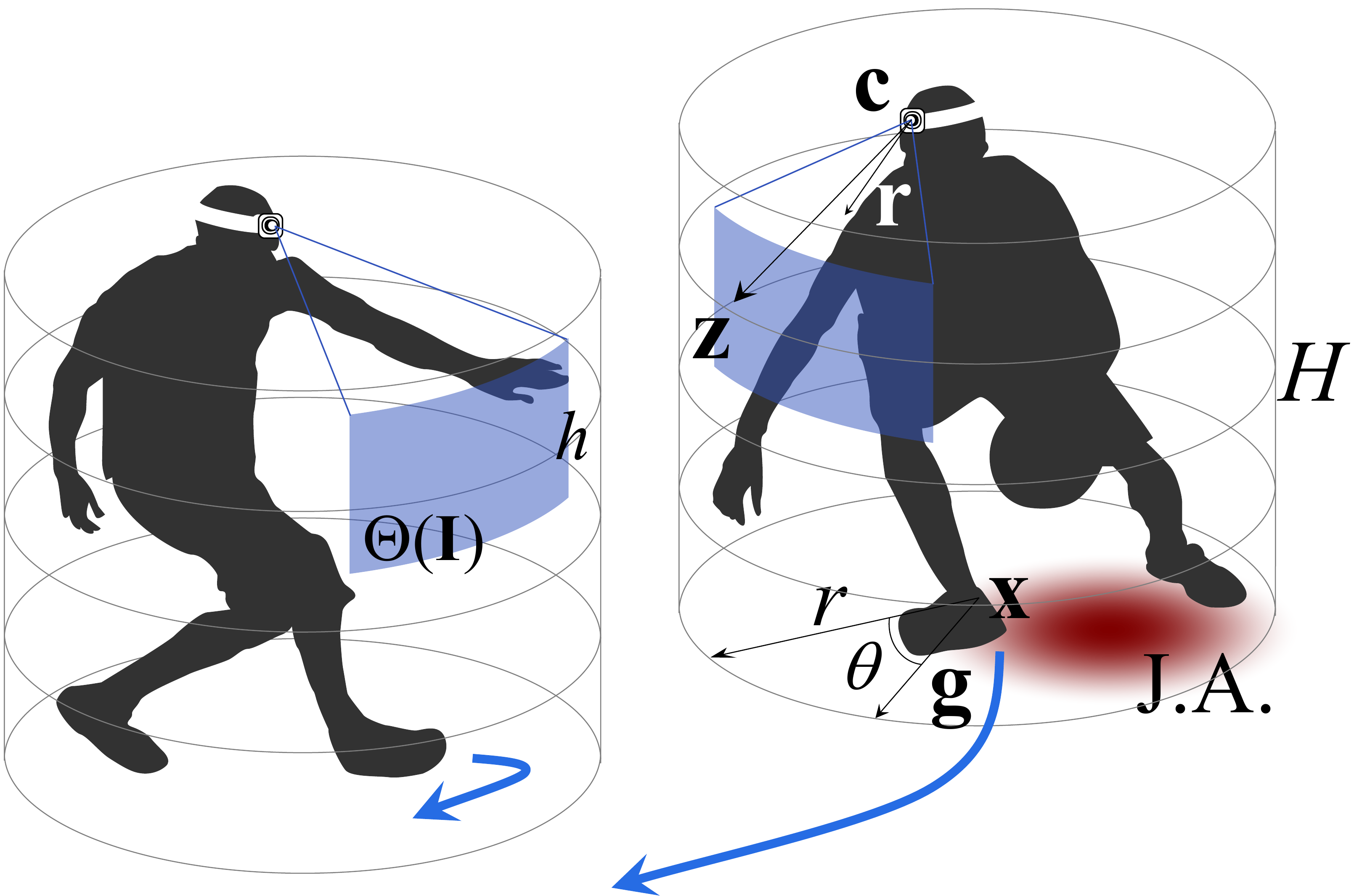}}~~ 
      \subfigure[Stabilized image]{\label{Fig:labeling}\includegraphics[height=0.137\textheight]{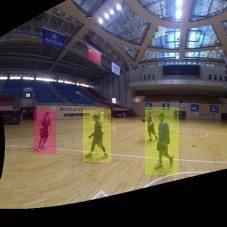}}~~
      \subfigure[Trajectory retrieval]{\label{Fig:retrieval}\includegraphics[height=0.137\textheight]{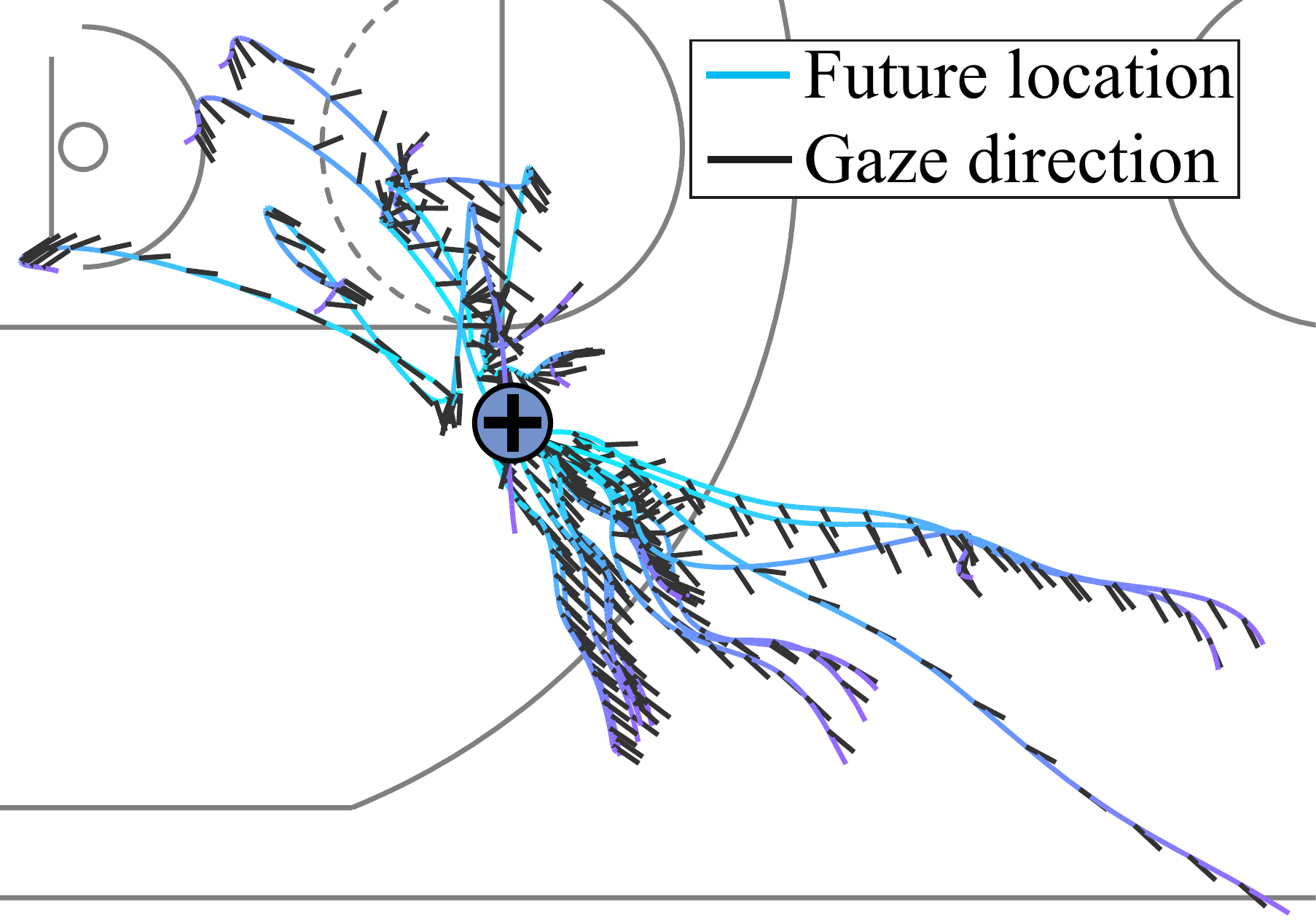}}~~
      \subfigure[Trajectory reprojection]{\label{Fig:reprojeciton}\includegraphics[height=0.137\textheight]{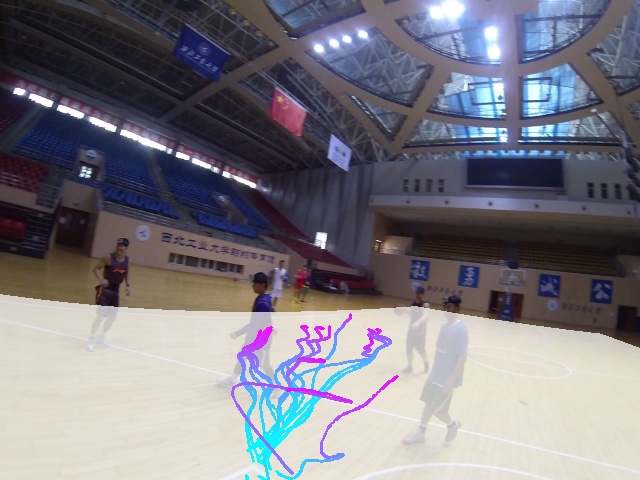}}
  \caption{(a) We model the space around a person using a cylinder that (b) stabilizes a first person image, $\Theta(\mathbf{I})$. The location and orientation of other players in the image are fully automatically labeled using 3D reconstruction. (c) We retrieve egocentric trajectories by associating the visual semantics of social configuration. (d) The retrieved trajectories are projected onto the first person image.} 
  \label{Fig:traj_retrieval}
\end{figure*}
Our work integrates two core vision tasks: 1) egocentric social perception: identifying social and spatial configuration, e.g., where I am, who I interact with, and how far they are, and 2) long term social behavior prediction: recognizing a plausible collective behaviors where we use joint attention as a social cue. 

Unlike third person vision systems operating in social scenes~\cite{cristani:2011,setti:2013,lan:2012,rodriguez:2011,choi:2014,chakraborty:2013,yang:2012,vignesh:2016}, a first person camera provides in-situ measurements of social interactions from an insider's perspective. This unique property allows a camera to record two sources of information simultaneously. (1) The 3D camera pose reconstructed by structure from motion approximates the gaze orientation, and the intersection of the gaze directions is the location of joint attention~\cite{park:2012,park:2013}. (2) The visual semantics (depth, edge, and surface) of first person images encodes what is socially salient. Faces have been used to recognize a group of people~\cite{fathi:2012} and build visual words to describe joint attention~\cite{pusiol:2014}. Subtle reciprocal behaviors can also be recognized~\cite{yonetani:2016}. Such visual information from first person cameras has been used for social video editing~\cite{arev:2014}, video summarization~\cite{lee:2012}, human-robot interactions~\cite{gori:2016}, and studying autistic behaviors for children~\cite{rehg:2013}.

How are my behaviors affected by others? This question has been a central theme in social psychology~\cite{allport:1985} and neuroscience, e.g., mirror neuron~\cite{rizzolatti:2004}, and their models inspire computational algorithms for multi-agent motion planning in robotics~\cite{cohen:2016,lee:2015,bhattacharya:2010} and graphics~\cite{karamouzas:2014,narrain:2009,pettre:2014}. A notable model is Helbing's social force model~\cite{helbing:1995} that explains crowd movements as a collection of physical interactions between social agents. This model is used to track a crowd~\cite{ali:2008} and recognize abnormal behaviors~\cite{mehran:2009}. 

A group as a whole naturally creates a distinctive geometry of social formation that accommodates its social activity, e.g., a street busker's performance surrounded by a crowd with a half circular formation. Therefore, the formation can be a key indicator to classify the type of social configurations that influence individual behaviors with respect to the group. For instance, Kendon's F-formation theory~\cite{kendon:1990} characterizes the spatial arrangements of a social group, that can be used to identify social interactions in an image~\cite{cristani:2011}, and its validity is empirically proven using a large social interaction dataset~\cite{park:2015}. In dynamic social scenes, the formation enables re-identifying a group of people in a crowd from non-overlapping camera views~\cite{alahi:2014}, and the progression of formation change can be learned via inverse reinforcement learning~\cite{ma:2016} and discriminative analysis (LSTM)~\cite{alahi:2016}. 

Note that most prior methods in predicting social behaviors rely on the third person measurements which have a limited access to how we perceive the social configurations. We leverage the visual social semantics embedded in first person cameras, which allows us to directly predict a plausible future group trajectory. This also enables predicting not only people's dynamic locations but also their attention, which have not been explored in prior studies.

\section{First Person Social Behavior Prediction}
We predict a group future trajectory (location and gaze direction) up to 5 seconds given their first person videos.
We use the 3D pose of a first person camera as a proxy of the head location, $\mathbf{c}$, and orientation (gaze direction)\footnote{Optionally, the fixed spatial relationship between camera optical axis and primary gaze direction can be calibrated~\cite{park:2012}.}, $\mathbf{r}$ where $\mathbf{c}$ and $\mathbf{r}$ are camera optical center and the $z$ axis of the camera rotation (optical axis) in the camera projection matrix, respectively. The camera projection matrices for all players are computed by structure from motion. We represent all variables in 2D by projecting 3D camera pose and joint attention on the 2D basketball court (50 ft.$\times$94 ft.) as shown in Figure~\ref{Fig:concept}: player's location $\mathbf{x}=\mathbf{c}_{1:2}$, gaze direction $\mathbf{g}=\mathbf{r}_{1:2}/\|\mathbf{r}_{1:2}\| \in \mathds{S}$, velocity $\mathbf{v}$, joint attention $\mathbf{s} \in \mathds{R}^2$ where $\mathbf{r}_{1:2}$ is the first two elements of $\mathbf{r}$ assuming that the coordinate system is aligned with the basketball court origin, i.e., $\left[\begin{array}{ccc} 0 & 0 & 1\end{array}\right]^\mathsf{T}$ is the surface normal of the court.
The ground truth joint attention is computed by triangulating the gaze directions of players~\cite{park:2012,park:2015}. For each player, a first person image, $\mathbf{I}$ is associated with the gaze, $(\mathbf{x},\mathbf{g})$. 

Our method is composed of two parts: 1) egocentric trajectory retrieval per player and 2) a group trajectory selection using a measure of social compatibility. For each player, we recognize images that have similar social and spatial configurations and retrieve a set of $N$ future trajectories (location and orientation) in Section~\ref{Sec:social_affordance}. This generates $nN$ trajectories for $n$ players, and we find a plausible group trajectory set that maximizes a measure of social compatibility (Section~\ref{Sec:compatibility}) while localizing joint attention using LRCN (Section~\ref{Sec:tracking}).




\begin{figure*}[th]
  \centering
  \subfigure[Visual selectivity]{\label{Fig:selectivity}\includegraphics[height=0.175\textheight]{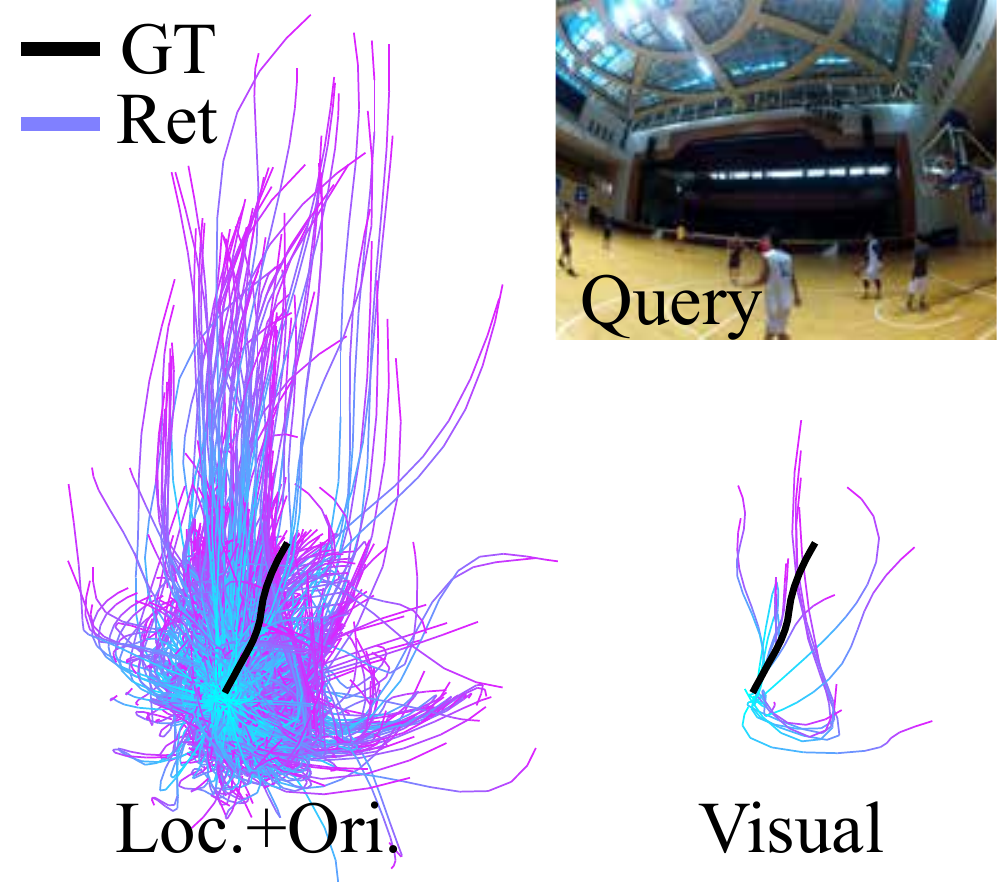}}~~~~
      \subfigure[Trajectory selection]{\label{Fig:concept}\includegraphics[height=0.175\textheight]{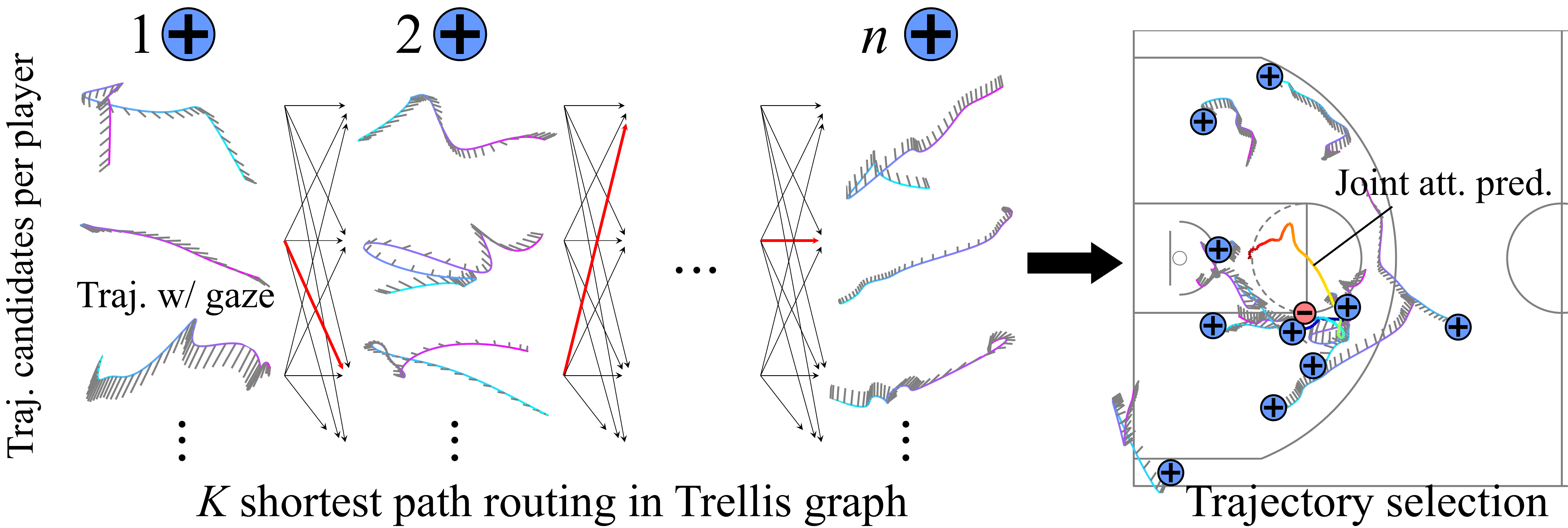}}
  \caption{(a) In conjunction with location and orientation prior, we use the visual semantics in a first person image to retrieve trajectories, which shows strong selectivity. GT: ground truth trajectory, Ret: retrieved trajectory. (b) We select $K$ best trajectory sets in a trellis graph using a generalized Dijkstra algorithm. A vertical space represents the retrieved trajectories per player and the path cost is computed by Equation~\ref{Eq:cost}. } 
  \label{Fig:traj}
\end{figure*}

\subsection{First Person Trajectory Retrieval} \label{Sec:social_affordance}
We behave similarly in similar social situations. The location, velocity, and orientation of other players are recorded in a first person image, $\mathcal{I}$, which encodes not only spatial layout, e.g., basket, center line, and background, but also social layout, e.g., where are other players, around the person. In this section, we learn the visual representation of social and spatial configurations from first person images.

We use the 3D reconstruction of first person videos to automatically annotate each other's location and orientation in pixels. We model each player using a cylinder with radius $r$ and height $H$ and project the cylinder onto a first person image, $\mathbf{I}$. The relative gaze direction, $\Delta \mathbf{g}$ is recorded in the label image, $\mathbf{M}$ using the HSV color map (Figure~\ref{Fig:labeling}):
\begin{align}
&\mathbf{M}_{xy} = \left\{\begin{array}{ll}
(0,0,0) & {\rm if}~~~ j^* = \emptyset\\
\left(\Delta\mathbf{g}_{j^*},0.9, 0.9\right) & {\rm otherwise}\end{array}\right. \nonumber\\
&{\rm where}~~~j^* = \underset{j}{\operatorname{argmin}}~\underset{\lambda}{\operatorname{min}} \{\lambda|\mathbf{c} + \lambda \mathbf{r} \in \mathcal{C}_j, \lambda > 0\},\nonumber
\end{align}
$\Delta\mathbf{g}=\angle\mathbf{g}_j-\angle\mathbf{g}$ is the $j^{\rm th}$ relative gaze direction, $\mathcal{C}_j = \{\mathbf{c}|\Delta\mathbf{c}_{3}<H, \|\Delta\mathbf{c}_{1:2}\| < r, \Delta\mathbf{c} = \mathbf{c}_j-\mathbf{c}\}$ is a set of 3D points in the cylinder of the $j^{\rm th}$ player. This label image directly encodes social configuration around a person. 

We stabilize a first person image onto the cylindrical surface\footnote{Similar projection has been used to generate a panoramic image~\cite{szeliski:1997}.} (Figure~\ref{Fig:labeling}). 
\begin{align}
&\Theta(\mathbf{I})_{\theta h} = \mathbf{I}_{xy}~~~~~~{\rm where}~~\left\{\begin{array}{l}x = \mathbf{r}_x^\mathsf{T} \mathbf{z}/\mathbf{r}^\mathsf{T} \mathbf{z}\\y = \mathbf{r}_y^\mathsf{T} \mathbf{z}/\mathbf{r}^\mathsf{T} \mathbf{z} \end{array}\right.\nonumber,
\end{align}
and $\mathbf{z} = \left[\begin{array}{ccc}\cos\theta & \sin\theta & h\end{array}\right]^\mathsf{T}$. $\mathbf{r}_x$ and $\mathbf{r}_y$ is the $X$ and $Y$ axes of the first person camera. The mapping function $\Theta$ applies to both first person image $\mathbf{I}$ and label image $\mathbf{M}$. 

The warped image has three properties that make visual learning effective. 1) Aligned vanishing lines: the head and foot location of the players are dependent solely on the depth given similar height; 2) no perspective distortion: the scale in image linearly proportional to the inverse depth; 3) optical center invariance: the representation is linear in angle where the optical center shift is linear translation in angle. 

We learn the visual social semantics using a Siamese neural network. We generate the positive and negative pairs of images based on $\mathbf{M}$, i.e., positive if $\|\Theta(\mathbf{M}_i)-\Theta(\mathbf{M}_j)\| < \epsilon$ and negative otherwise. We minimize the following contrastive loss for training:
\begin{align}
L_{\rm soc} = \sum_{(i,j)\in\mathcal{P}} l_{ij} \|\Delta \phi \|^2+(1-l_{ij}) \max(0, m^2-\|\Delta \phi \|^2)  \nonumber
\end{align}
where $l_{ij}$ is a label indicating positive and negative pairs, $\phi(\Theta(\mathbf{I}))$ is the visual feature of the warped image $\Theta(\mathbf{I})$ learned by a convolutional neural network (CNN). $\Delta\phi = \phi(\Theta(\mathbf{I}_i))-\phi(\Theta(\mathbf{I}_j))$, $\mathcal{P}$ is the set of pairs, and $m$ is a margin between positive and negative pairs. We use the pre-trained CNN~\cite{krizhevsky:2012} and refine the weights through the training.  

We empirically observed that this pairing across all locations inclines to learn the background because a first person image is dominated by background pixels, e.g., the network learns ego-motion rather than social configurations~\cite{jayaraman:2015,agrawal:2015,jayaraman:2016,park_floc:2016}. Instead, we make pairs that are located and oriented in the similar area of the basketball courts, i.e., $\|\mathbf{x}_i-\mathbf{x}_j\| < \epsilon_\mathbf{x}$ and $|\angle \mathbf{g}_i-\angle\mathbf{g}_j| < \epsilon_\mathbf{g}$. Our learning based approach is beneficial in particular dynamic social scenes that include severe motion blur, illumination and view point changes where standard structure from motion often fails.

Based on the learned feature of the target image, $\phi(\Theta(\mathbf{I}_{\rm tar}))$, we retrieve $N$ 2D trajectories, $\mathcal{T} = \{\mathbf{T}|\epsilon > \|\phi(\Theta(\mathbf{I}_{\rm tar}))-\phi(\Theta(\mathbf{I}))\|\}$ where $\mathbf{T}=\left\{\mathbf{x}^t,\mathbf{g}^t\right\}_{t=1}^T$ is a trajectory (location and gaze direction) of each player and $\epsilon$ is the feature decision boundary learned by the neural network. Similar to the training phase, we restrict the training data samples based on location and orientation. In practice, we cluster the trajectories, $\mathbf{T}$ using Medoidshift~\cite{sheikh:2007} to identify topologically distinctive trajectories~\cite{park_floc:2016}. Figure~\ref{Fig:retrieval} illustrates the retrieved trajectories that are projected onto the first person image in Figure~\ref{Fig:reprojeciton}. Note that our first person trajectory retrieval is highly selective as shown in Figure~\ref{Fig:selectivity}.

The retrieved trajectories have three properties: 1) they discover egocentric physical space to move based on social configurations; 2) they include diverse topological structure, i.e., different trajectories may be plausible given a social configuration; and 3) they reflect spatial layout.

\begin{figure*}[th]
  \centering  
      \subfigure[Joint attention prediction]{\label{Fig:ja_prediction}\includegraphics[height=0.18\textheight]{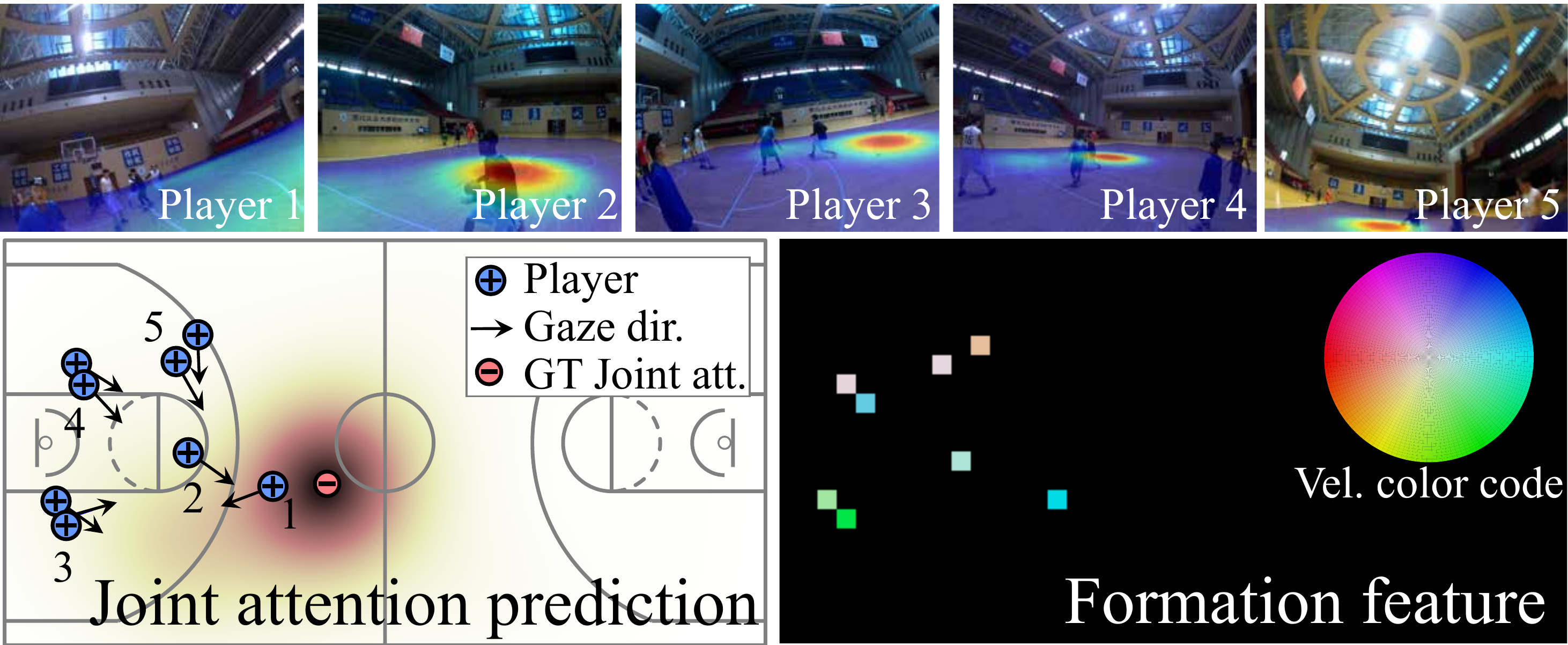}}~~
        \subfigure[Dynamics of joint attention]{\label{Fig:lstm}\includegraphics[height=0.18\textheight]{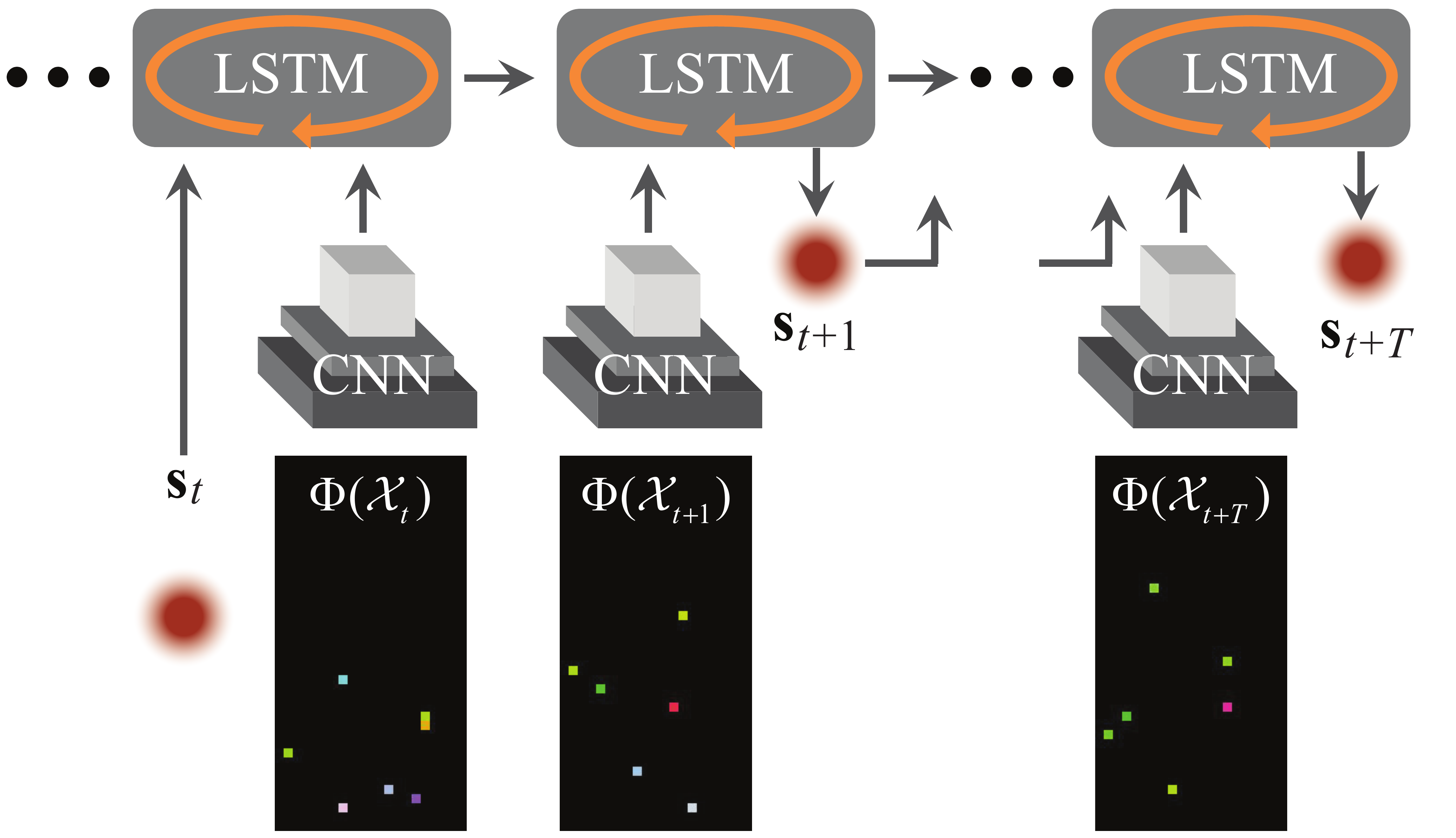}}~
  \caption{(a) We predict joint attention using a social formation feature that encodes the player's location and instantaneous velocity (bottom right). Top row: the predicted joint attention is projected onto each first person image. (b) The social formation features are used to learn the dynamics of joint attention using LRCN~\cite{donahue:2015}.} 
  \label{Fig:vel}
\end{figure*}

\subsection{Group Trajectory via Social Compatibility} \label{Sec:compatibility}

There exist $N^n$ possible combinations of group trajectories where $n$ is the number of players. The trajectories are retrieved independently, and not all combinations are {\em socially} plausible. In this section, we recognize the plausible trajectory combinations using a measure of social compatibility---the gaze alignment towards joint attention predicted by social formation. Note that we consolidate all retrieved egocentric trajectories by registering them into the basketball court.


There are two ways of computing joint attention in a static social scene: 1) geometrically finding the intersection of gaze directions~\cite{park:2012} and 2) statistically learning the characteristics of the social formation, which does not require knowing gaze directions~\cite{park:2015}. Note that we denote the geometrically computed joint attention as $\mathbf{s}$ to differentiate with the statistically estimated joint attention $\hat{\mathbf{s}}$.

Ideally, these two locations of joint attention must agree, and we define a measure of social compatibility based on the alignment between two joint attentions:
\begin{align}
\eta = \frac{1}{n} \sum_{\mathbf{x} \in \mathcal{X}} \frac{\left(\hat{\mathbf{s}} - \mathbf{x}_i\right)^\mathsf{T}\mathbf{g}_i}{\|\hat{\mathbf{s}} - \mathbf{x}_i\|}, \nonumber
\end{align}
where $\mathcal{X} = \{\mathbf{x}_i, \mathbf{g}_i\}_{i=1}^n$ is a set of player locations and gaze directions. The social compatibility measures how the gaze directions are geometrically aligned with statistically computed joint attention, and it characterizes which social formation and corresponding gaze directions are socially plausible. Note that $\hat{\mathbf{s}}$ is a function of $\{\mathbf{x}_i\}_{i=1}^n$.


We integrate the social compatibility over time to evaluate a group trajectory set:
\begin{align}
\eta = \frac{1}{nT} \sum_{i=1}^n \eta_t \left(\{\hat{\mathbf{s}}_t\}_{t=1}^T, \mathbf{T}_i\right),\nonumber
\end{align}
where $\eta_t$ is the accumulated measure of social compatibility over $T$ time instances.

\noindent\textbf{Group Trajectory Selection} Among the $nN$ retrieved trajectories from all players, $\left\{\mathcal{T}_i\right\}_{i=1}^n$, 
we find a group trajectory set that maximizes the measure of social compatibility:
\begin{align}
\underset{\{p_i\}_{i=1}^n}{\operatorname{argmin}}~~-\frac{1}{nT}\sum_{i=1}^n \eta_t \left(\{\hat{\mathbf{s}}_t\}_{t=1}^T, \mathbf{T}_{p_i}\right), \label{Eq:cost}
\end{align}
where $\{p_i\}_{i=1}^n$ is an index set for the retrieved trajectory of each player.

Solving Equation~(\ref{Eq:cost}) by the exhaustive search is computationally prohibited, $\mathcal{O}(N^n)$. A stochastic search such as Monte Carlo simulations does not apply due to the low probability to choose a correct model. Instead, we employ the generalized Dijkstra algorithm, or Yen's algorithm~\cite{yen:1971} to efficiently find the $K$ best trajectory sets. 

We construct a trellis graph where the vertical slice represents a set of the retrieved trajectories per players, $\mathcal{T}_i$, i.e., each node is a trajectory and an edge indicates the trajectory selection as shown in Figure~\ref{Fig:traj}. A path along the trellis graph determines the selected trajectory set where the path cost is defined in Equation~(\ref{Eq:cost}). Despite the greedy search due to a nonlinear prediction of joint attention (Section~\ref{Sec:tracking}), in practice, the algorithm finds ``good'' solutions that have high social compatibility. We predict the group behaviors using the selected trajectory set.

\begin{figure*}[th]
  \centering  
      \subfigure[Joint attention vs. speed ]{\label{Fig:ja}\includegraphics[height=0.17\textheight]{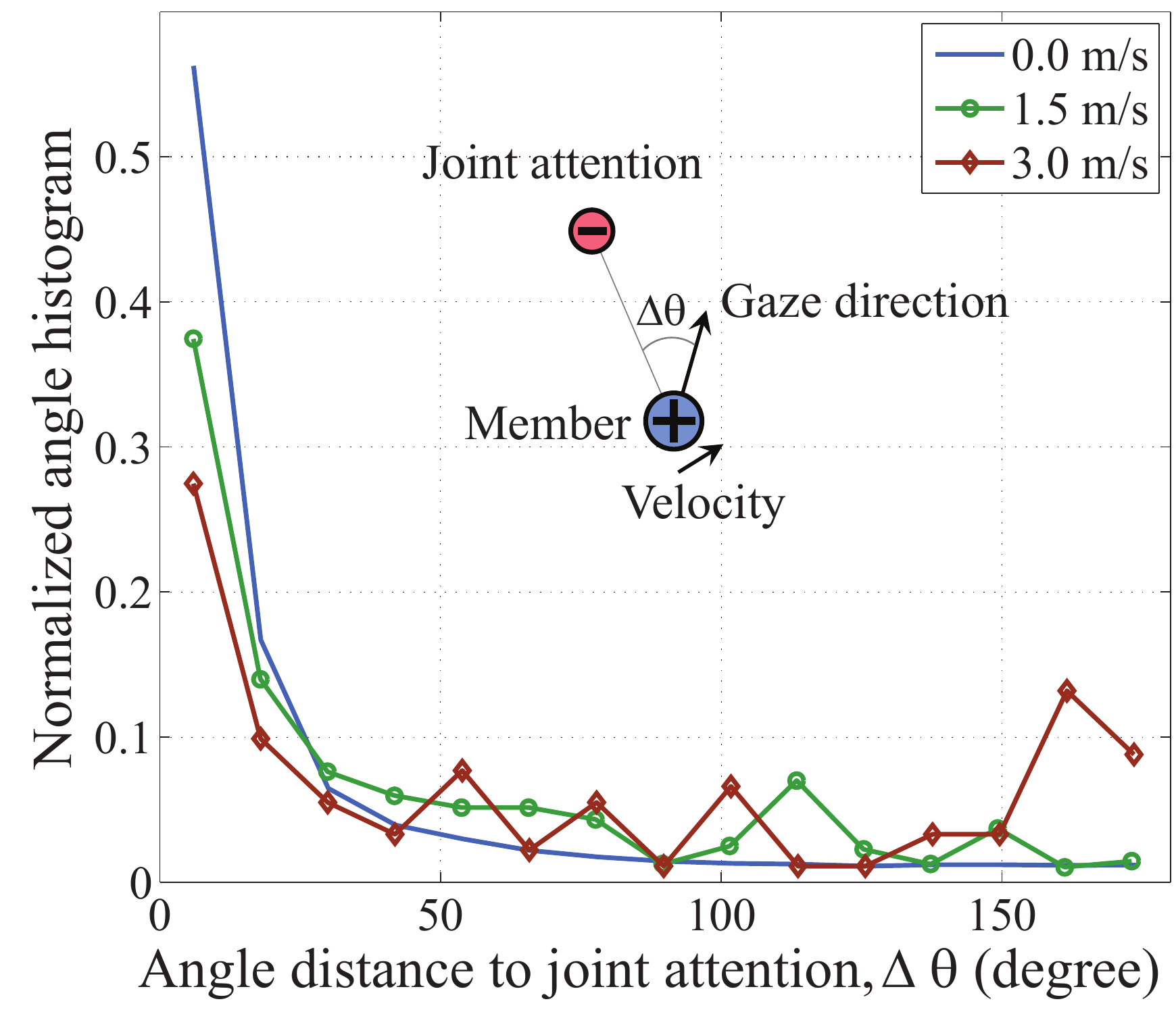}} ~~~~~~~~
      \subfigure[Spatial distribution of role]{\label{Fig:role1}\includegraphics[height=0.17\textheight]{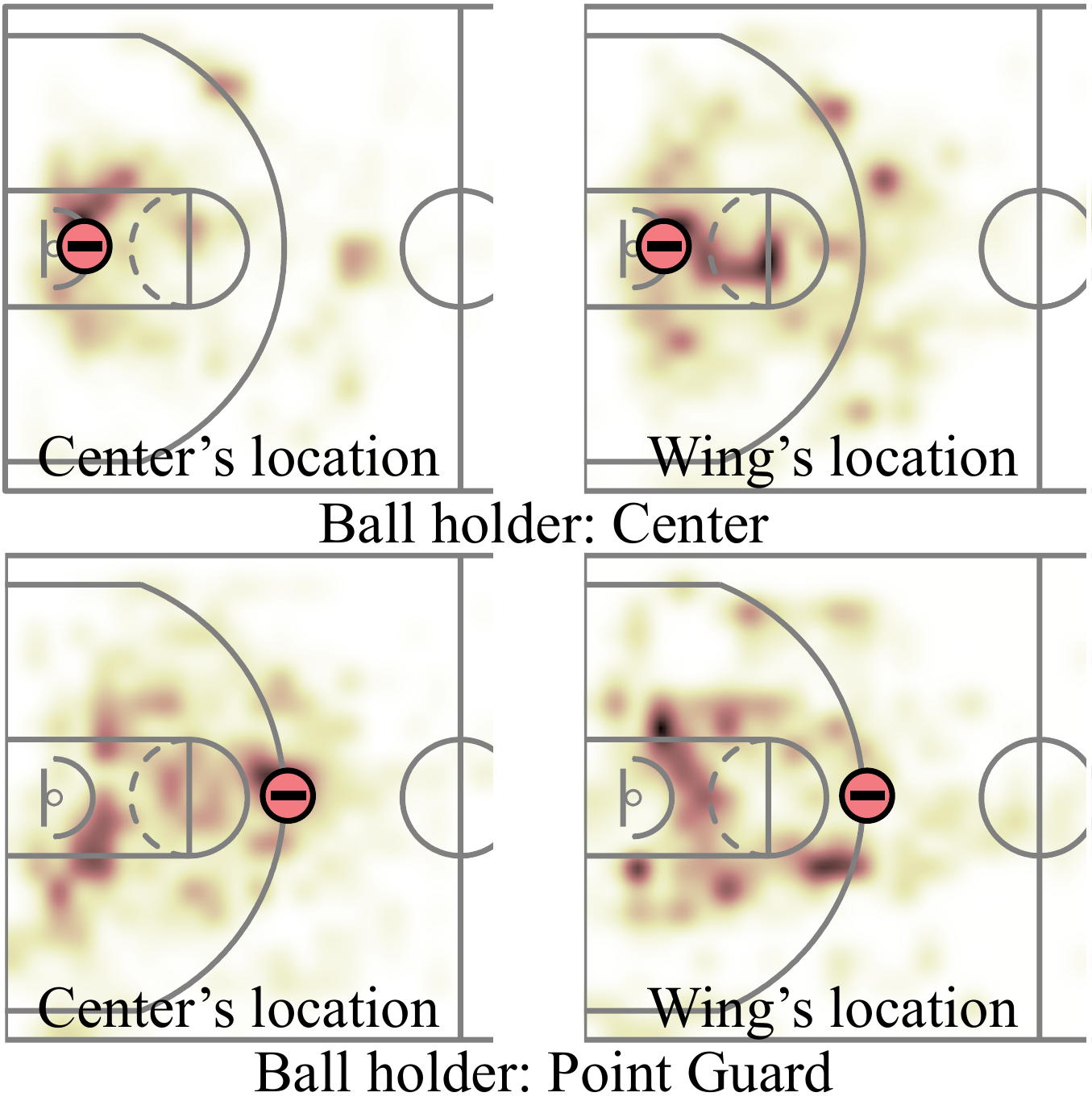}}  ~~~~~~~~
      \subfigure[Role correlation]{\label{Fig:role}\includegraphics[height=0.17\textheight]{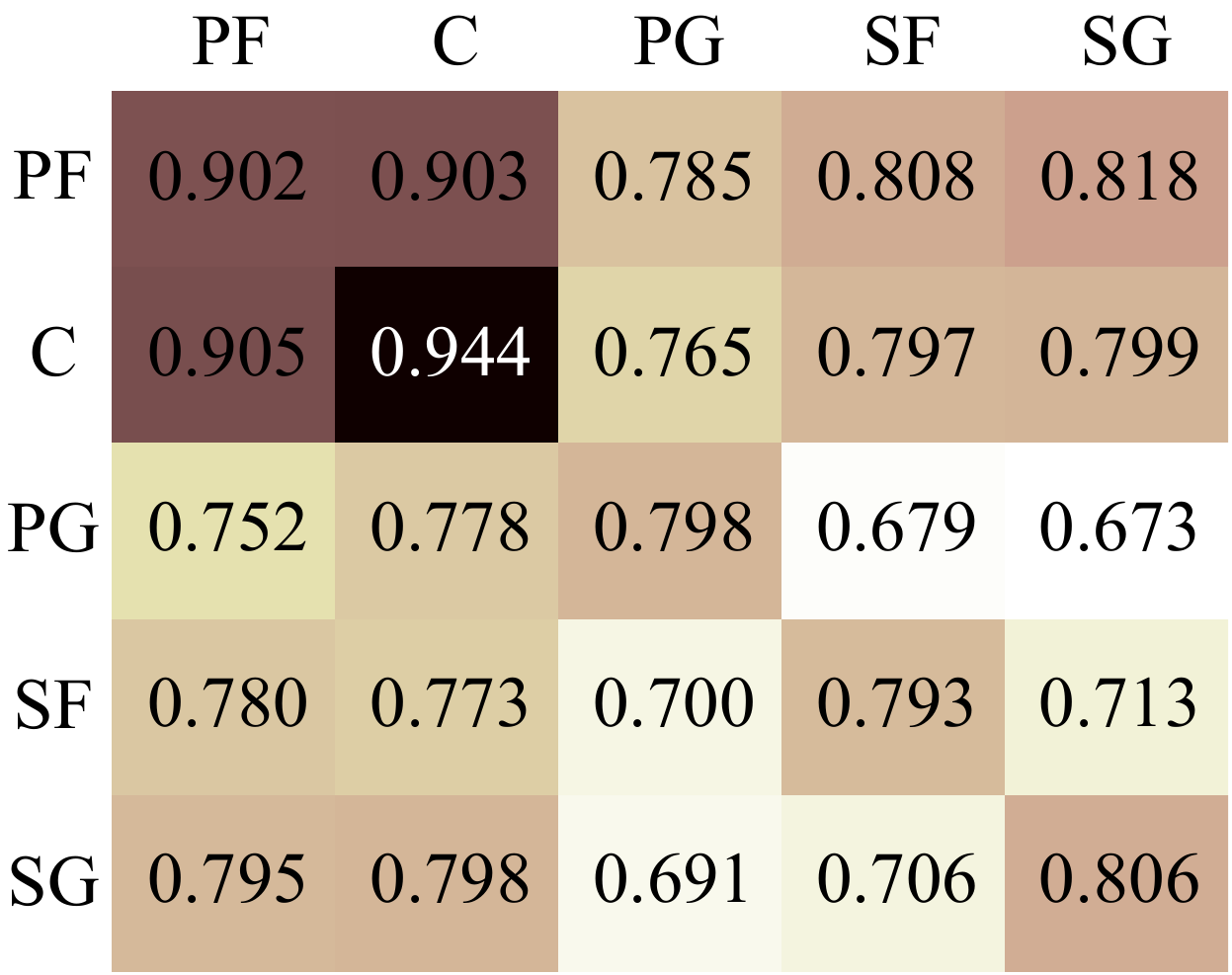}} 
  \caption{(a) Players consistently engage joint attention while playing. They look at the joint attention more than 60 \% of their play. (b) Role is a key factor to determine social formations. We illustrate distributions of the Center and Wing players given the ball holder's location.  (c) The role of a player is a strong prior to predict other players, e.g., two Centers from different teams often move together to block each other. PF: Power Forward, C: Center, PG: Point guard, SF: Small Forward, SG: Shooting guard.} 
  \label{Fig:vel}
\end{figure*}
\subsection{Joint Attention Dynamics} \label{Sec:tracking}
Equation~(\ref{Eq:cost}) requires joint attention prediction, $\hat{\mathbf{s}}$. In this section, we learn the dynamics of joint attention with respect to social formation using LRCN~\cite{donahue:2015}.

As an input of the network, we generate a formation feature image, $\Phi(\mathcal{X})$ that encodes the occupancy and instantaneous velocity, $\mathbf{v}$, of the players in a discretized basketball court. The HSV value of the formation feature image is set to: 
\begin{align}
&\Phi_{ij} (\mathcal{X}) = \left\{\begin{array}{ll}
\left(\angle\bar{\mathbf{v}}_{ij},\|\bar{\mathbf{v}}_{ij}\|, 0.9\right) & {\rm if}~~~ |\mathcal{C}_{ij}| > 0\\(0,0,0) & {\rm if}~~~ |\mathcal{C}_{ij}| = 0\end{array}\right. \nonumber\\
&{\rm where}~~~\bar{\mathbf{v}}_{ij} = \frac{1}{|\mathcal{C}_{ij}|} \sum_{\mathbf{v}_k \in \mathcal{C}_{ij}} \mathbf{v}_k,\nonumber
\end{align}
$\mathcal{C}_{ij} = \left\{\mathbf{v} | \mathbf{x} \in C_{ij}\right\}$ and the $(i,j)$ cell of the court. $\Phi(\mathcal{X})$ is illustrated in the bottom right of Figure~\ref{Fig:ja_prediction}. Note that unlike social dipole moment~\cite{park:2015}, this representation is independent on the location of center of mass and joint attention, which is robust to missing data.

We use LRCN with a few minor modifications to learn the dynamics of joint attention. We minimize the following joint attention error:
\begin{align}
    L_{\rm LSTM} = \sum_{t = 1}^T \|\mathbf{s}_t-\hat{\mathbf{s}}_t \|^2, \nonumber
\end{align}
where the $\hat{\mathbf{s}}_{t+1}$ is recursively computed by
\begin{align}
\hat{\mathbf{s}}_{t+1} = f(\hat{\mathbf{s}}_t,\Phi(\mathcal{X}_t);\mathbf{w}_{\rm CNN},\mathbf{w}_{\rm LSTM}). \label{Eq:dynamics}
\end{align}
$f$ is the dynamics parametrized by the weights of a convolutional neural network, $\mathbf{w}_{\rm CNN}$, and a long short-term memory unit, $\mathbf{w}_{\rm LSTM}$ as shown in Figure~\ref{Fig:lstm}. We initialize $\mathbf{w}_{\rm CNN}$ based on pre-trained model~\cite{krizhevsky:2012} separately with further refinement by regressing the static location of joint attention from social formation, $\mathbf{s} = g(\Phi(\mathbf{X});\mathbf{w}_{\rm CNN})$.

\section{Basketball Dataset Analysis}

We use the first person basketball video data collected by the university team at Northwestern Polytechnical University in China~\cite{arev:2014,park:2015}. The dataset includes 10.5 hours of basketball games. 
We take two steps for reconstruction: (1) reference reconstruction: we subsample images from each player to reconstruct the reference 3D points and cameras ($\sim$3,000 images) using structure from motion~\cite{hartley:2004}; and (2) camera registration: we register each image into the reference reconstruction coordinate system using a camera resectioning algorithm~\cite{lepetit:2008} with local bundle adjustment up to 500 consecutive images.  

Figure~\ref{Fig:ja} illustrates a normalized angle histogram of joint attention engagement. This indicates that the players consistently align their gaze directions to joint attention ($<$ 40 degree): 83\%, 65\%, and 48\% of their play at 0 m/s, 1.5 m/s, and 3 m/s speed, respectively. As the speed gets faster, the player's gaze direction tends to deviate from the joint attention: it often follows the fast motion, which forms behind the person (180 degrees) at high speed. 

Player's role is a key factor to characterize social formations. Figure~\ref{Fig:role1} illustrates a spatial distribution of players based on their role, given joint attention. For instance, when the Center possesses a ball, Power Forward and Center are likely located near the basket area for blocking and rebound. When a Point Guard possesses the ball, players tend to be distributed widely to create space to receive the ball. Also the role is a strong predictor of the play as similar roles in different teams enforces them to move together. Figure~\ref{Fig:role} shows that a strong correlation of roles in different teams. 






\section{Result}
\begin{figure}[t]
  \centering  
  \label{Fig:ja_quant}\includegraphics[width=0.45\textwidth]{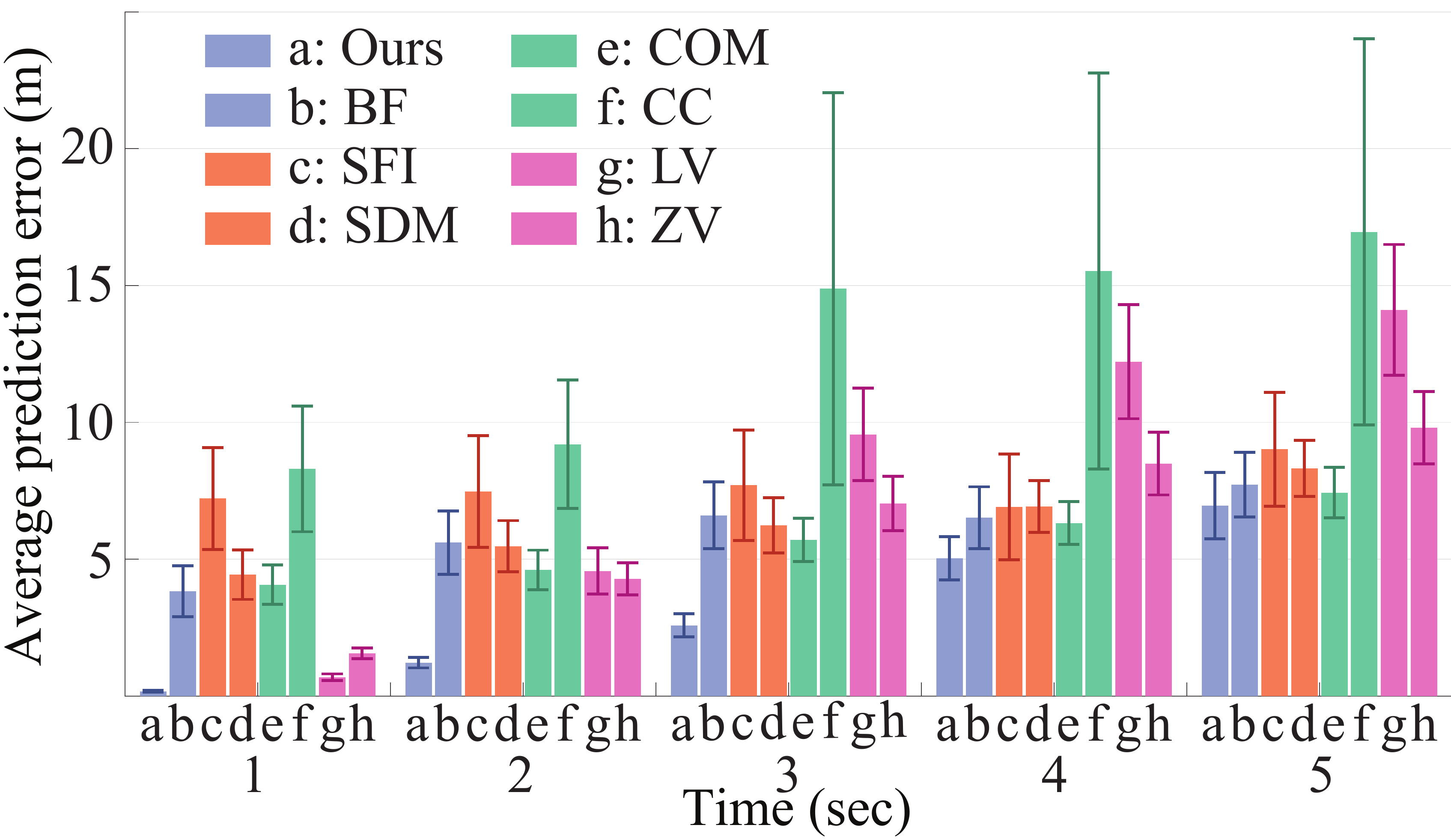}
  \caption{We compare our predicted joint attention with 7 baseline algorithms which it shows 7 m error after 5 seconds. See text for the baseline algorithms.} 
  \label{Fig:vel}
\end{figure}

\begin{figure*}[th]
  \centering  
  \subfigure[Missing data prediction]{\label{Fig:missing_data}\includegraphics[height=0.195\textheight]{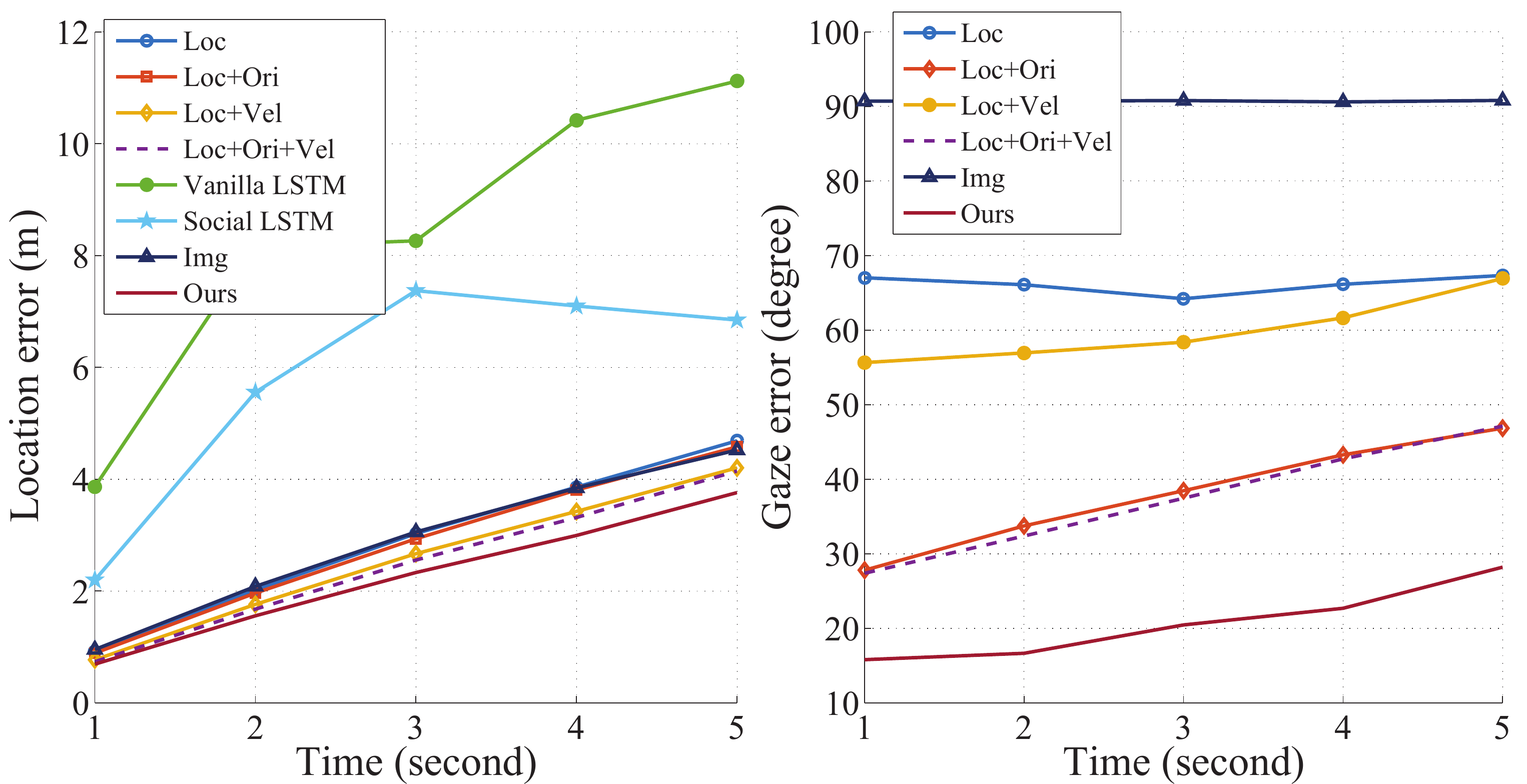} }~~~
      \subfigure[Group behavior prediction]{\label{Fig:group_prediction}\includegraphics[height=0.2\textheight]{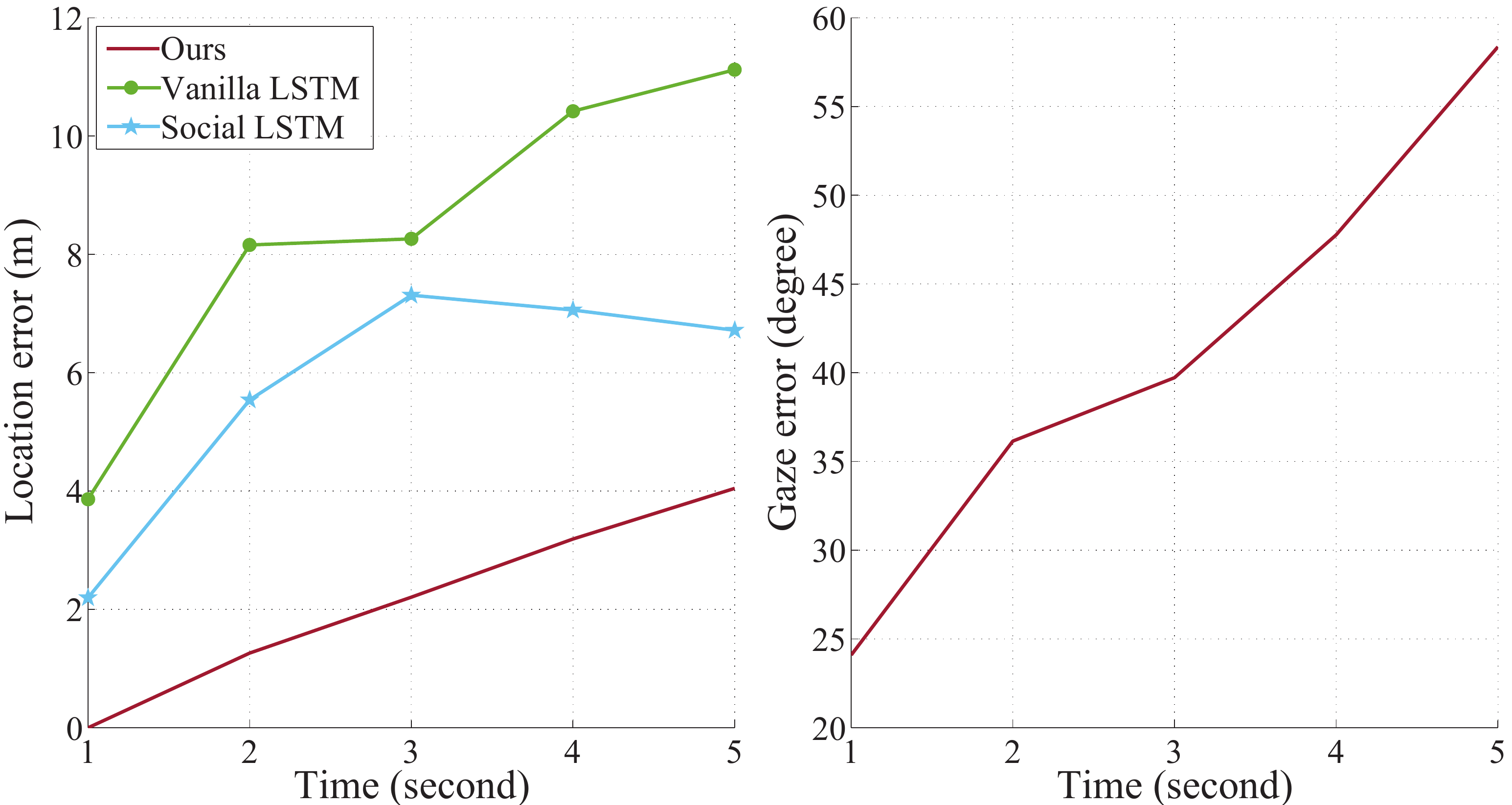} }
  \caption{(a) We evaluate our algorithm by comparing with 7 baseline algorithms including state-of-the-art Social LSTM and our method consistently outperforms other methods on missing data prediction task. In particular, our method shows a strong predictive power on gaze direction (30 degree after 5 seconds). (b) We predict a group trajectory set and compare with Vanilla LSTM and Social LSTM while no comparable algorithm exists for graze prediction.} 
  \label{Fig:vel}
\end{figure*}

We evaluate our social behavior prediction by comparing with the ground truth data. Note that the testing data are completely isolated from the training data in terms of time and players.  

We use AlexNet~\cite{krizhevsky:2012} to train the Siamese network with Caffe~\cite{jia2014caffe}. 240k image pairs are generated from first person images of players where the pairs are selected within similar location ($\epsilon_\mathbf{x} < $ 3m) and orientation ($\epsilon_\mathbf{g} <$ 45 degrees) in the basketball court. Due to the location and orientation prior, the network can be efficiently trained with strong generalization power (98.7\% testing accuracy). For training the dynamics of joint attention, we concatenate the AlexNet FC7 layer with LSTM through Theano~\cite{2016arXiv160502688short}. We generate 85k sequences of joint attention and corresponding social formation feature (210$\times$410). The testing average error over 5 seconds is 3.12m.

\subsection{Quantitative Evaluation} 
We evaluate our prediction in three categories: joint attention, missing trajectory, and social trajectories. 

\noindent\textbf{Joint attention prediction} We compare our method with 7 baseline algorithms for predicting 5 seconds. A) Zero velocity (ZV) and linear constant velocity (LV) extrapolate the location of joint attention by taking into account instantaneous velocity; B) Center of mass (COM) and center of circumcircle (CC) are geometrically computed based on the locations of players; C) Social dipole moment~\cite{park:2015} (SDM) is used to learn a binary classifier (AdaBoost) to recognize the location of joint attention; D) Our social formation feature image (SFI), $\Phi(\mathcal{X})$, with LSTM is used to predict joint attention using a convolutional neural network~\cite{krizhevsky:2012}. We train the network to minimize the Euclidean loss of $\|\mathbf{s}-\hat{\mathbf{s}}\|$; E) A Bayesian filtering (BF) is applied for temporal smoothing by learning a stochastic dynamics of joint. 

Figure~\ref{Fig:ja_quant} illustrates the predictive validity where our method outperforms all baseline algorithms. In particular, it shows a strong predictive power up to 4 seconds with 5 m error in a highly dynamic scene. The error in LV and ZV indicates the nature of dynamics of the basketball game. COM, CC, SDM, and SFI are time independent predictors where COM shows the most consistent and strongest prediction. This is caused by the fact that social formations in basketball data are often distributed near the basket area where the center of mass of players is likely located.

\noindent\textbf{Missing trajectory prediction} We apply our method for missing trajectory prediction. We leave out a trajectory and predict its behaviors using social compatibility. 


We compare our method with 7 baseline algorithms. A) We use a kinematic prior to predict a trajectory: location (Loc), orientation (Ori), velocity (Vel), and their combinations. B) We compare with state-of-the-art third person prediction systems based on Vanilla LSTM~\cite{greff:2015} and Social LSTM~\cite{alahi:2016}. We use the occupancy based Social LSTM which applies pooling based on social proximity. C) We compare with first person prediction based solely on visual features (Img) (no kinematic knowledge). The visual features are learned by our Siamese network. Note that we compare not only future locations but also gaze directions except for Vanilla and Social LSTMs where gaze prediction is not possible with their trivial extension.

Figure~\ref{Fig:missing_data} indicates that orientation or velocity is a strong prior to predict future while our method produces more selective trajectories due to the social compatibility measure. Vanilla LSTM produces unconvincing results due to its limited expressibility on social interactions and Social LSTM shows drifts because the behaviors of basketball players are often affected by long range team players. Notably a first person image based method without kinematic knowledge (Img) performs poorly, which indicates visual information alone can be ambiguous.

Our method outperforms all baseline algorithms. In particular, our method shows strong predictive power on gaze direction driven by joint attention (30 degree error after 5 seconds).

\noindent\textbf{Social trajectory prediction} We focus on comparing with third person approaches: Vanilla LSTM and Social LSTM. Note that both LSTMs require longer observation time (10 seconds) to predict 5 seconds while our first person based method needs 0.5 second (instantaneous velocity).

Note that Vanilla LSTM behaves similarly to the missing data prediction as it has no consideration on social behaviors. Our method produces the error range, 5 m and 30 degree error after 5 seconds as shown in Figure~\ref{Fig:group_prediction}.  

We also characterize the prediction error based on player's role summarized in Table~\ref{table:role}. This error indicates that the predictive power can differ by the roles, e.g., predicting Shooting Guard's behaviors is relatively more difficult than Centers because they involve with diverse interactions across the court.

\subsection{Qualitative Evaluation}
\begin{figure*}[th]
  \centering  
  \subfigure[Taking-turn]{\label{Fig:missing_data}\includegraphics[width=\textwidth]{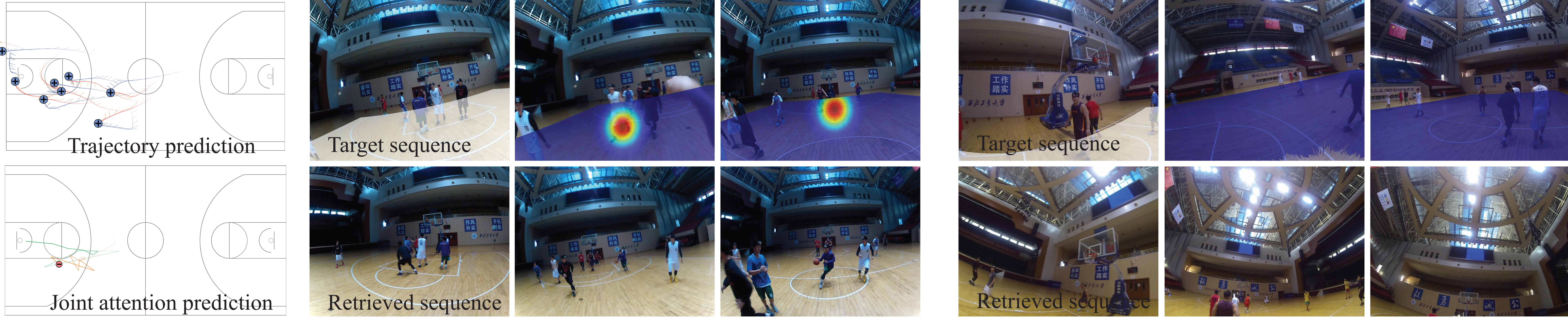} }
  \subfigure[Attack]{\label{Fig:missing_data}\includegraphics[width=\textwidth]{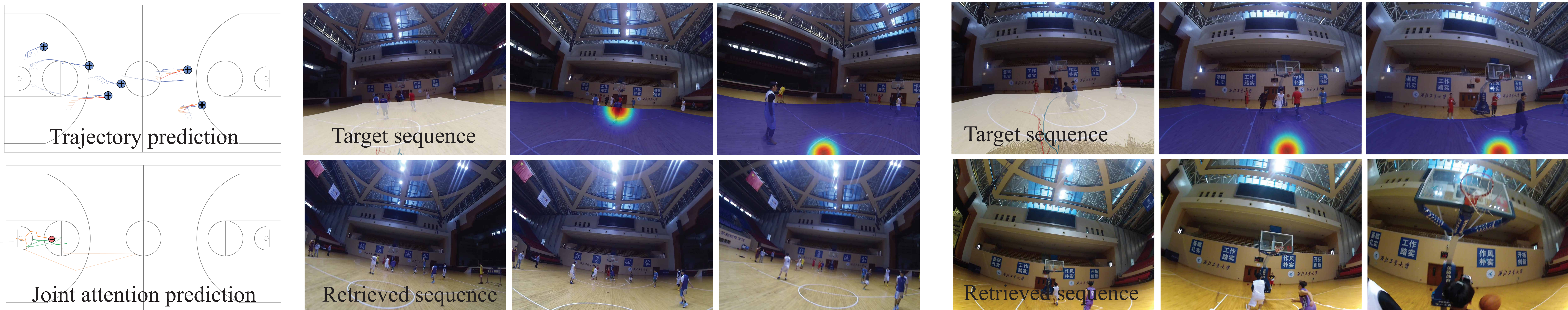} }
  
  \subfigure[Drive-in]{\label{Fig:missing_data}\includegraphics[width=\textwidth]{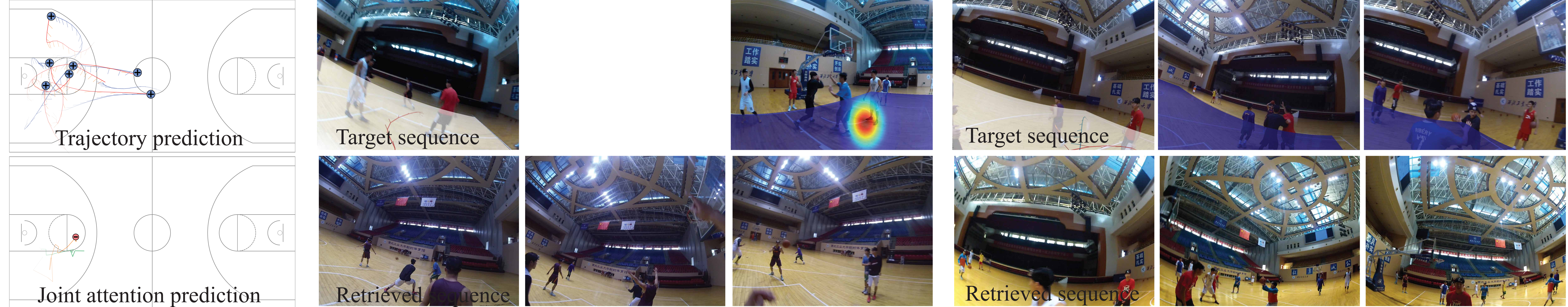} }
  \subfigure[Shot]{\label{Fig:missing_data}\includegraphics[width=\textwidth]{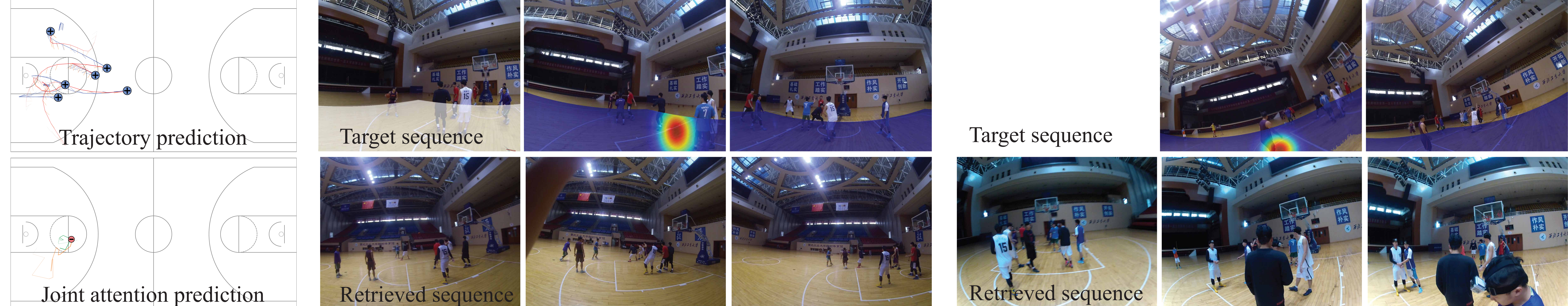} }
  \caption{We evaluate our algorithm qualitatively in diverse scenarios (taking-turn, attack, drive-in, and shot). The first column and top row: a comparison between the predicted trajectories with gaze directions in blue with ground truth trajectory in red up to 5 seconds. First column and bottom row: a comparison between the predicted joint attention in green with the ground truth joint attention in orange. Transparency encodes time. Second column: a comparison between a target sequence (top row) and the retrieved sequence (bottom row). We also show the retrieved sequences to reason about our prediction. The retrieved sequence has similar social configuration as time evolves. The predicted trajectories and joint attention are projected onto the target sequence to validate the prediction. The joint attention agrees with scene activities. The blank space is missing data where structure from motion fails.}
  \label{Fig:qual}
\end{figure*}

We apply our method to predict players future behaviors in diverse basketball scenarios. Figure~\ref{Fig:qual} shows trajectory and joint attention predictions. We also show the retrieved sequences that have similar social configuration to reason about predictions. 

\begin{figure}[h]
  \centering  
  \label{Fig:average}\includegraphics[width=0.48\textwidth]{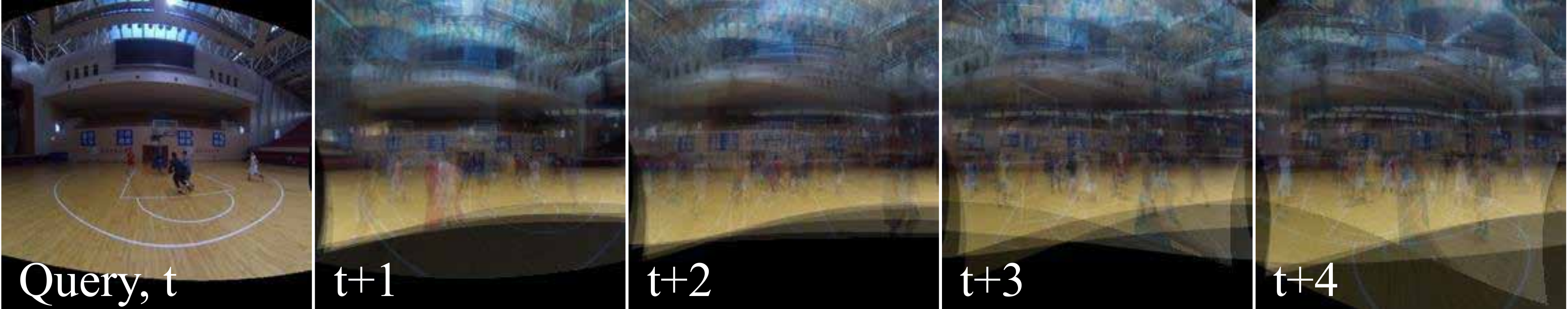}
  \caption{We predict future images based on our $K$ best solutions.} 
\end{figure}
\vspace{-3mm}
\noindent\textbf{Predicting future image} The capability of predicting gaze direction enables hallucinating future images, e.g., what would I see in a next few seconds? Figure~\ref{Fig:average} visualizes the average future images retrieved by the $K$ best solutions. The average image is aligned with background structure and social configurations while it starts to dissolve as time progresses.

\begin{table*}[th]
\centering
\footnotesize
\begin{tabular}{l|c|c|c||c|c|c||c|c|c||c|c|c||c|c|c}
\hline
 & \multicolumn{3}{c||}{Power Forward}& \multicolumn{3}{c||}{Point Guard}& \multicolumn{3}{c||}{Small Forward}& \multicolumn{3}{c||}{Shooting Guard}& \multicolumn{3}{c}{Center}\\
 \cline{2-16}
 & 1sec & 3 sec & 5 sec& 1sec & 3 sec & 5 sec& 1sec & 3 sec & 5 sec& 1sec & 3 sec & 5 sec& 1sec & 3 sec & 5 sec \\\hline
Ours &  \textbf{0.50}  & \textbf{0.60} & \textbf{0.40} & \textbf{1.59} &  \textbf{3.76} & \textbf{5.71} & \textbf{0.64} &  \textbf{0.25} & \textbf{2.41} & \textbf{1.51} &  \textbf{4.95} & \textbf{7.79} & \textbf{1.39} &  \textbf{0.65} & \textbf{1.70}\\
Vanilla LSTM & 6.50 & 10.86 & 13.77 & 6.85 & 12.86 & 8.30 & 3.52 & 3.81 &  9.05 & 1.98 & 15.43 &  7.69 & 9.35 &  5.54 & 12.23\\
Social LSTM & 2.98 & 3.02 & 3.59 & 6.55 & 11.49 & 12.32 & 0.53 & 5.35 & 7.84 &  1.99 & 10.19& 2.43 & 5.14&  1.60 & 7.36\\
\hline
\end{tabular}
\caption[Trajectory prediction error based on player's role]{Trajectory prediction error based on player's role}
\label{table:role}
\end{table*}

\section{Summary}
We present a method to predict the future location and gaze direction of basketball players from their first person videos. 3D reconstruction of multiple first person videos provides the automatic supervision for learning visual social semantics. We use the learned representation to retrieve trajectories per player. We evaluate the plausibility of each group trajectory using social compatibility. We select $K$ best group trajectories using a generalized Dijkstra's algorithm. We demonstrate that our first person based method is effective, outperforming state-of-the-art social activity prediction systems that use third person views.


\newpage
{\small
\bibliographystyle{ieee}
\bibliography{egbib}
}

\end{document}